# Probabilistic Deep Learning with Probabilistic Neural Networks and Deep Probabilistic Models

Daniel T. Chang (张遵)

*IBM (Retired)* dtchang43@gmail.com

**Abstract:**   Probabilistic deep learning is deep learning that accounts for uncertainty, both model uncertainty and data uncertainty. It is based on the use of probabilistic models and deep neural networks. We distinguish two approaches to probabilistic deep learning: probabilistic neural networks and deep probabilistic models. The former employs deep neural networks that utilize probabilistic layers which can represent and process uncertainty; the latter uses probabilistic models that incorporate deep neural network components which capture complex non-linear stochastic relationships between the random variables. We discuss some major examples of each approach including Bayesian neural networks and mixture density networks (for probabilistic neural networks), and variational autoencoders, deep Gaussian processes and deep mixed effects models (for deep probabilistic models). TensorFlow Probability is a library for probabilistic modeling and inference which can be used for both approaches of probabilistic deep learning. We include its code examples for illustration.

## 1 Introduction

Standard deep learning is deterministic and does not account for *uncertainty*, either model uncertainty or data uncertainty [1]. This is a major limitation, particularly for mission-critical and real-world applications such as biomedical applications. *Probabilistic deep learning* removes this limitation by accounting for uncertainty, both model uncertainty and data uncertainty.

Probabilistic deep learning is based on the use of *probabilistic models* and *deep neural networks*. We distinguish two approaches to probabilistic deep learning: *probabilistic neural networks* and *deep probabilistic models*, due to their difference in focus. The former employs deep neural networks that utilize *probabilistic layers* [7] which can represent and process uncertainty; the latter uses probabilistic models that incorporate *deep neural network components* [10] which capture complex non-linear stochastic relationships between the random variables. The distinction, however, may not always be clear cut: A probabilistic neural network may be used to realize a deep probabilistic model, and a deep probabilistic model may incorporate probabilistic neural network components.

For probabilistic neural networks, we discuss two major examples: Bayesian neural networks and mixture density networks. *Bayesian neural networks* [8] utilize probabilistic layers that capture *uncertainty over weights and activations*, and are trained using *Bayesian inference*. *Mixture density networks* [9] are designed to quantify neural network *prediction uncertainty*. They consist of two components: a mixture model and a deep neural network.

For deep probabilistic models, we discuss three major examples: variational autoencoders, deep Gaussian processes, and deep mixed effects models. *Variational autoencoders* [1] are prescribed deep generative models realized using an autoencoder neural network that consists of an *encoder network* and a *decoder network*, which encodes a data sample to a *latent representation* and generates data samples from the latent space, respectively. *Deep Gaussian processes* [11] are multi-layer generalizations of Gaussian processes [2], which capture the *compositional nature of (deterministic) deep learning* while mitigating some of the disadvantages through a *Bayesian approach*. *Deep mixed effects models* [12] are extensions to linear mixed effects models that use *nonlinear mappings* between the response variable and predictor variables, which are realized using *deep neural networks*.

*TensorFlow Probability* [4-6] is a library for probabilistic modeling and inference which, with Keras / TensorFlow, can be used for building both probabilistic neural networks and deep probabilistic models. We include its code examples, where applicable, for illustration.

## 2 Background Information

As background information, we briefly discuss uncertainty, probabilistic models, and Bayesian inference, which are the probabilistic foundation of probabilistic deep learning. We also briefly summarize TensorFlow Probability. The discussions on uncertainty and probabilistic models are extracted from [1]. Please see it for references and further discussions.

### 2.1 Uncertainty

*Uncertainty* arises because of limitations in our ability to observe the world, limitations in our ability to model it, and possibly inherent non-determinism (e.g., quantum phenomena). From the perspective of deep learning, there are two types of uncertainty: model uncertainty and data uncertainty. *Model uncertainty* accounts for uncertainty in the model structure and model parameters. This is important to consider for mission-critical applications and small datasets. *Data uncertainty* accounts for out-of-distribution data and noisy data. This is important to consider for real-world applications and large datasets.

Uncertainty, therefore, should be an inescapable aspect of deep learning. Because of this, we need to allow our learning system to represent uncertainty and our reasoning system to consider different possibilities due to uncertainty. Further, to obtain meaningful conclusions, we need to reason not just about what is possible, but also about what is probable. *Probability* *theory* provides us with a formal framework for representing uncertainty and for considering multiple possible outcomes and their likelihood.

## 2.2 Probabilistic Models

The probabilistic approach to modeling uses probability theory to represent all forms of uncertainty and for inference:

- *Probability distributions* are used to represent all uncertain elements in a model (including structural, parametric, out-of-distribution and noise-related) and how they relate to the data.

- The basic *rules of probability theory* are used to infer the uncertain elements given the observed data.

- Learning from data occurs through the transformation of the *prior distributions* (defined before observing the data) into *posterior distributions* (obtained after observing the data). The application of probability theory to learning from data is called *Bayesian learning* or *Bayesian inference*.

Simple probability distributions over a single or a few *random variables* can be composed to form the building blocks of larger, more complex models. The compositionality of probabilistic models means that the behavior of these building blocks in the context of the larger model is often much easier to understand. The dominant paradigm for representing such *compositional probabilistic models* is probabilistic graphical models.

### *Probabilistic Graphical Models*

*Probabilistic graphical models (PGMs)* are structured probabilistic models. A PGM is a way of describing a probability distribution of random variables, using a *graph* to describe which variables in the probability distribution directly interact with each other, with each *node* representing a variable and each *edge* representing a direct interaction. This allows the models to have significantly fewer parameters which can in turn be estimated reliably from less data. These smaller models also have dramatically reduced computational cost in terms of storing the model, performing inference in the model, and drawing samples from the model.

Usually a PGM comes with both a *graphical representation* of the model and a *generative process* to depict how the random variables are generated. Due to its Bayesian nature, a PGM is easy to extend to incorporate other information or to perform other tasks. There are essentially two types of PGM: directed PGM (also known as *Bayesian network*) and undirected PGM (also known as *Markov random field*).

A good PGM needs to accurately capture the distribution over the *observed variables* $x$, p($x$). Often the different elements of $x$ are highly dependent on each other. The approach most commonly used to model these dependencies is to introduce several *latent variables* $z$. The model can then capture dependencies between any pair of variables $x_i$ and $x_j$ indirectly, via direct dependencies between $x_i$ and $z$, and direct dependencies between $z$ and $x_j$. Latent variables have advantages beyond their role in efficiently capturing p($x$). The latent variables $z$ also provides an alternative representation for $x$. Many approaches accomplish *representation learning* by learning latent variables.

## 2.3 Bayesian Inference

The main idea of *Bayesian inference* [3] is to infer a *posterior distribution* over the parameters $\theta$ of a probabilistic model given some observed data D using Bayes theorem as:

$$p(\theta \mid D) = p(D \mid \theta)p(\theta) / p(D) = p(D \mid \theta)p(\theta) / \int p(D \mid \theta)p(\theta)d\theta,$$

where p(D | $\theta$) is the *likelihood*, p(D) is the *marginal likelihood* (or *evidence*), and p($\theta$) is the *prior*. Computing the posterior, and moreover sampling from it, is usually intractable, especially since computing the evidence (integrals) is hard. *Approximate inference* methods are usually used, including:

1. Markov Chain Monte Carlo: Approximates integrals via sampling.
2. Variational Inference: Approximates integrals via optimization.

The Bayesian posterior can be used to model new unseen data D∗ using the *posterior predictive*:

$$p(D^* \mid D) = \int p(D^* \mid \theta)p(\theta \mid D)d\theta,$$

which is also called the *Bayesian model average* because it averages the predictions of all plausible models weighted by their posterior probability.

The *choice of prior* [3], p($\theta$), is one of the most critical parts of the Bayesian inference. It should be chosen in a way such that it accurately reflects our beliefs about the parameters $\theta$ before seeing any data. There is no universally preferred prior, but each probabilistic model / inference task is potentially endowed with its own optimal prior.

## 2.4 TensorFlow Probability

*TensorFlow Probability (TFP)* [4-6] is a library for probabilistic modeling and inference in TensorFlow. It provides integration of probabilistic models with deep neural networks, gradient-based inference via automatic differentiation, and scalability to large datasets and models via hardware acceleration (e.g., GPUs) and distributed computation.

TFP is structured as following layers, from bottom to top:

1. TensorFlow: Numerical operations.
2. Probabilistic building blocks
    a. Distributions (*tfp.distributions*): A large collection of probability distributions which provide fast, numerically stable methods for generating samples and computing statistics.
    b. Bijectors (*tfp.bijectors*): Reversible and composable transformations of distributions, with automatic caching, preserving the ability to generate samples and compute statistics.
3. Probabilistic model / neural network building
    a. Joint Distributions (*tfp.distributions.JointDistributionSequential* and others): Joint distributions over one or more possibly-interdependent distributions, designed for building small- to medium-size PGMs.
    b. Probabilistic Layers (*tfp.layers*): Neural network layers with uncertainty.
4. Bayesian inference
    a. Markov Chain Monte Carlo (*tfp.mcmc*): Algorithms for approximating integrals via sampling.
    b. Variational Inference (*tfp.vi*): Algorithms for approximating integrals via optimization.
    c. Optimizers (*tfp.optimizer*): Stochastic optimization methods.
    d. Monte Carlo (*tfp.monte_carlo*): Tools for computing Monte Carlo expectations.

TFP can be used for building both probabilistic neural networks and deep probabilistic models, as can be seen in subsequent discussions.

## 3 Probabilistic Neural Networks

*Probabilistic neural networks* are deep neural networks that utilize probabilistic layers which can represent and process uncertainty. *Probabilistic layers* [7] can capture uncertainty over weights (Bayesian neural networks), pre-activation units (dropout), activations ("probabilistic output layers"), or the function itself (Gaussian processes). They can also be layers that propagate uncertainty from input to output. Probabilistic layers are designed to be drop-in replacement of their deterministic counter parts.

## 3.1 Bayesian Neural Networks

*Bayesian neural networks (BNNs)* [8] utilize probabilistic layers that capture *uncertainty over weights and activations*, and are trained using *Bayesian inference*. At a high level, they can be represented as follow:

$$\theta \sim p(\theta),$$

$$y = BNN_\theta(x) + \epsilon,$$

where $\theta$ represents BNN parameters and $\epsilon$ represents random noise.

To design a BNN, the first step is to choose a neural network architecture, e.g., a convolutional neural network. Then, one has to choose a probability model: a *prior distribution* over the possible model parameters $p(\theta)$ and a *prior confidence* in the predictive power of the model $p(y \mid x, \theta)$, i.e., $BNN_\theta(x)$. For supervised learning, the *Bayesian posterior* can be written as:

$$p(\theta \mid D) = p(D_y \mid D_x, \theta)p(\theta) / \int p(D_y \mid D_x, \theta)p(\theta)d\theta,$$

where D is the training set, $D_x$ the training features, and $D_y$ the training labels. Computing the posterior, and moreover sampling from it, is usually intractable, especially since computing the evidence (denominator) is hard. In general one uses a *variational inference* approach, which learns a variational distribution $q_\phi(\theta)$ to approximate the exact posterior.

Given the Bayesian posterior, or in practice its variational approximation, the *posterior predictive* can be computed as:

$$p(y \mid x, D) = \int p(y \mid x, \theta)p(\theta \mid D)d\theta.$$

In practice, the distribution $p(y \mid x, \theta)$ is sampled indirectly using $BNN_\theta(x)$, and $\theta$ is sampled from the variational distribution $q_\phi(\theta)$.

The following shows an example code of a convolutional *BNN model in TensorFlow Probability* [4]. It defines a LeNet-5 model using three convolutional (with max pooling) layers and two fully connected dense layers. It uses the *Flipout Monte Carlo estimator* for these layers. *KL divergence*, using Lambda function, is passed as input to the kernel_divergence_fn on flipout layers. The Keras API will automatically add the KL divergence to the cross entropy loss, effectively calculating the (negated) Evidence Lower Bound Loss (ELBO).

```
kl_divergence_function = (lambda q, p, _: tfd.kl_divergence(q, p) /
                             tf.cast(NUM_TRAIN_EXAMPLES, dtype=tf.float32))

model = tf.keras.models.Sequential([
    tfp.layers.Convolution2DFlipout(
        6, kernel_size=5, padding='SAME',
        kernel_divergence_fn=kl_divergence_function,
        activation=tf.nn.relu),
    tf.keras.layers.MaxPooling2D(
        pool_size=[2, 2], strides=[2, 2],
        padding='SAME'),
    tfp.layers.Convolution2DFlipout(
        16, kernel_size=5, padding='SAME',
        kernel_divergence_fn=kl_divergence_function,
        activation=tf.nn.relu),
    tf.keras.layers.MaxPooling2D(
        pool_size=[2, 2], strides=[2, 2],
        padding='SAME'),
    tfp.layers.Convolution2DFlipout(
        120, kernel_size=5, padding='SAME',
        kernel_divergence_fn=kl_divergence_function,
        activation=tf.nn.relu),
    tf.keras.layers.Flatten(),
    tfp.layers.DenseFlipout(
        84, kernel_divergence_fn=kl_divergence_function,
        activation=tf.nn.relu),
    tfp.layers.DenseFlipout(
        NUM_CLASSES, kernel_divergence_fn=kl_divergence_function,
        activation=tf.nn.softmax)
])
```

### 3.2 Mixture Density Networks

*Mixture density networks (MDNs)* [9] are designed to quantify neural network *prediction uncertainty*. MDNs consist of two components: a mixture model and a deep neural network. The *mixture model* can be written as:

$$p(\mathbf{y} \mid \mathbf{x}) = \sum_{k=1}^{K} p(\mathbf{y}; \phi_k(\mathbf{x})) p(\phi_k(\mathbf{x}); \pi(\mathbf{x})).$$

The deep neural network can be of any valid architecture that maps **x** onto both the parameters $\{\phi_k(\mathbf{x})\}_{k=1}^{K}$ of the K *mixture components* and the parameters $\pi(\mathbf{x})$ of the *mixing distribution*. MDNs can be trained by maximizing the *log-likelihood* of the parameters of the deep neural network given the training set $D = \{\mathbf{x}_n, \mathbf{y}_n\}_{n=1}^{N}$ using gradient-based optimizers.

The following shows an example code of a *MDN model in TensorFlow Probability* [4]. It utilizes a mixture distribution layer with independent normal (Gaussian) components.

```
event_shape = [1]
num_components = 5
params_size = tfpl.MixtureNormal.params_size(num_components, event_shape)

model = tfk.Sequential([
  tfkl.Dense(12, activation='relu'),
  tfkl.Dense(params_size, activation=None),
  tfpl.MixtureNormal(num_components, event_shape)
])
```

# 4 Deep Probabilistic Models

*Deep probabilistic models* [10] are probabilistic models that incorporate *deep neural network components* which capture complex non-linear stochastic relationships between the random variables.

*Unsupervised and semi-supervised learning* methods are often based on *generative models* which are probabilistic models that express hypotheses about the way in which data may have been generated. *PGMs* have emerged as a broadly useful approach to specifying generative models. *Deep generative models (DGMs)* [1] are PGMs that employ *deep neural networks* for parameterizing the models. In particular, *prescribed DGMs* are those that provide an explicit parametric specification of the probability distribution of the observed variable **x**, specifying the *likelihood function* $p_\theta(x)$ with parameter **θ**. DGMs are among the most widely used deep probabilistic models.

## 4.1 Variational Autoencoders

(The discussion in this subsection is extracted from [1]. Please see it for references and further discussions.)

The *variational autoencoder (VAE)* is a prescribed DGM realized using an autoencoder neural network that consists of an *encoder network* and a *decoder network*, which encodes a data sample to a *latent representation* and generates data samples from the latent space, respectively. The decoder network is a differentiable generator network and the encoder network is an auxiliary inference network. Both networks are jointly trained using variational learning. *Variational learning* uses the *variational lower bound* of the marginal log-likelihood as the single objective function to optimize both the generator network and the auxiliary inference network.

The likelihood function of a prescribed DGM can be written as:

$$p_\theta(x) \equiv p_\theta(z)\, p_\theta(x \mid z)$$

It is usually intractable to directly evaluate and maximize the marginal log-likelihood $\log(p_\theta(x))$. Following the *variational inference* approach, one introduces an *auxiliary inference model* $q_\varphi(z \mid x)$ with parameters $\varphi$, which serves as an approximation to the exact posterior $p_\theta(z \mid x)$.

The *variational lower bound* can be rewritten as:

$$L(x;\, \theta,\, \varphi) = \sum_z q_\varphi(z \mid x) \log(p_\theta(x \mid z)) - D_{KL}(q_\varphi(z \mid x) \,\|\, p_\theta(z))$$

The first term is the expected reconstruction quality, requiring that $p_\theta(x \mid z)$ is high for samples of $z$ from $q_\varphi(z \mid x)$, while the second term (the KL divergence between the approximate posterior and the prior) acts as a regularizer, ensuring we can generate realistic data by sampling latent variables from $p_\theta(z)$.

*Variational learning* is to maximize the variational lower bound over the training data. It performs something like the autoencoder, with $q_\varphi(z \mid x)$ *as the encoder* and $p_\theta(x \mid z)$ *as the decoder*. As such, the VAE introduces the constraint on the autoencoder that the latent variable $z$ is distributed according to a *prior p(z)*. The encoder $q_\varphi(z \mid x)$ approximates the *posterior p(z | x)*, and the decoder $p_\theta(x \mid z)$ parameterizes the *likelihood p(x | z)*. The *generation model* is then $z \sim p(z);\ x \sim p(x \mid z)$.

The following shows an example code of a convolutional *VAE in TensorFlow Probability* [4]: encoder, decoder and VAE. Note that KLDivergenceRegulizer is used in the encoder.

```python
encoder = tfk.Sequential([
    tfkl.InputLayer(input_shape=input_shape),
    tfkl.Lambda(lambda x: tf.cast(x, tf.float32) - 0.5),
    tfkl.Conv2D(base_depth, 5, strides=1,
                padding='same', activation=tf.nn.leaky_relu),
    tfkl.Conv2D(base_depth, 5, strides=2,
                padding='same', activation=tf.nn.leaky_relu),
    tfkl.Conv2D(2 * base_depth, 5, strides=1,
                padding='same', activation=tf.nn.leaky_relu),
    tfkl.Conv2D(2 * base_depth, 5, strides=2,
                padding='same', activation=tf.nn.leaky_relu),
    tfkl.Conv2D(4 * encoded_size, 7, strides=1,
                padding='valid', activation=tf.nn.leaky_relu),
    tfkl.Flatten(),
    tfkl.Dense(tfpl.MultivariateNormalTriL.params_size(encoded_size),
               activation=None),
    tfpl.MultivariateNormalTriL(
        encoded_size,
        activity_regularizer=tfpl.KLDivergenceRegularizer(prior)),
])
```

```python
decoder = tfk.Sequential([
    tfkl.InputLayer(input_shape=[encoded_size]),
    tfkl.Reshape([1, 1, encoded_size]),
    tfkl.Conv2DTranspose(2 * base_depth, 7, strides=1,
                         padding='valid', activation=tf.nn.leaky_relu),
    tfkl.Conv2DTranspose(2 * base_depth, 5, strides=1,
                         padding='same', activation=tf.nn.leaky_relu),
    tfkl.Conv2DTranspose(2 * base_depth, 5, strides=2,
                         padding='same', activation=tf.nn.leaky_relu),
    tfkl.Conv2DTranspose(base_depth, 5, strides=1,
                         padding='same', activation=tf.nn.leaky_relu),
    tfkl.Conv2DTranspose(base_depth, 5, strides=2,
                         padding='same', activation=tf.nn.leaky_relu),
    tfkl.Conv2DTranspose(base_depth, 5, strides=1,
                         padding='same', activation=tf.nn.leaky_relu),
    tfkl.Conv2D(filters=1, kernel_size=5, strides=1,
                padding='same', activation=None),
    tfkl.Flatten(),
    tfpl.IndependentBernoulli(input_shape, tfd.Bernoulli.logits),
])
```

```
vae = tfk.Model(inputs=encoder.inputs,
                outputs=decoder(encoder.outputs[0]))
```

## 4.2 Deep Gaussian Processes

*Gaussian Processes*

(The discussion in this subsection is extracted from [2]. Please see it for references and further discussions.)

The *Gaussian process (GP)* is a well-known *non-parametric* probabilistic model. A Gaussian process is a generalization of the *Gaussian distribution*. Whereas a probability distribution describes random variables which are scalars or vectors, a *stochastic process* governs the properties of *functions*. We can loosely think of a function as *an infinite vector*, each entry in the vector specifying the function value $f(x_i)$ at a particular input $x_i$. A key aspect of this is that if we ask only for the properties of the function at *a finite number of points*, then *GP inference* will give us the same answer if we ignore the infinitely many other points.

In the *function-space view* a Gaussian process defines a *distribution over functions*, and inference takes place directly in the *space of functions*. A Gaussian process is completely specified by its *mean function* (average of all functions) and *kernel (covariance) function* (how much individual functions can vary around the mean function):

$m(\mathbf{x}) = \mathbb{E}[f(\mathbf{x})],$

$k(\mathbf{x}, \mathbf{x'}) = \mathbb{E}[(f(\mathbf{x}) - m(\mathbf{x}))(f(\mathbf{x'}) - m(\mathbf{x'}))]$

As such, we write the Gaussian process as

$f(\mathbf{x}) \sim \mathcal{GP}(m(\mathbf{x}), k(\mathbf{x}, \mathbf{x'})).$

Under the Gaussian process, the true objective is modeled by a *GP prior* with a mean and a kernel function. Given a set of (noisy) *observations* from initial evaluations, a Bayesian posterior update gives the *GP posterior* with an *updated mean and kernel function*. The mean function of the GP posterior gives the best *predictions* at any point conditional on the available observations, and the kernel function quantifies the *uncertainty* in the predictions. The *GP prior* is a multivariate Gaussian distribution:

$$p(\mathbf{f} \mid \boldsymbol{\theta}) \sim \mathcal{N}(\boldsymbol{\mu}, \mathbf{K}(\boldsymbol{\theta})).$$

So is the GP posterior:

$$p(\mathbf{f} \mid \mathcal{D}, \boldsymbol{\theta}) \sim \mathcal{N}(\boldsymbol{\mu'}, \mathbf{K'}(\boldsymbol{\theta})).$$

In the above, $\boldsymbol{\theta}$ is *GP parameters* (parameters of the GP kernel function).

*Deep Gaussian Processes*

Gaussian processes have the following limitations [11]:

- The assumption of *Gaussian marginals*: A Gaussian process cannot model heavy-tailed, asymmetric or multimodal marginal distributions.
- The requirement to define a prior entirely in terms of *mean and covariance*: Even very complicated covariance functions cannot express prior relationships that are not captured by mean and covariance alone.
- The *predictive covariance* of the exact posterior process is *independent of the observed outputs*: It is unsatisfactory that the predictive uncertainty is decoupled from the observations.
- The *predictive mean is linear* in the observations: There are many situations where this linearity might be inappropriate.

*Deep Gaussian processes (DGPs)* [11] are multi-layer generalizations of GPs that promise to overcome some of the above limitations of the single layer model. The DGP presents a paradigm for building deep models from a Bayesian perspective. It captures the *compositional nature of (deterministic) deep learning* while mitigating some of the disadvantages, e.g., lack of treatment for uncertainty, through a *Bayesian approach*.

Independent GPs $f^1, \ldots, f^L$ (at layer $l = 1, \ldots, L$) can be combined through *function composition* to construct a DGP [11]. The composite function $g(x)$ is given by

$$g(x) = f^L(f^{L-1}(\ldots f^1(x))),$$

where the outputs of each layer are the inputs to the next higher layer. Given a likelihood $p(y \mid g(x))$, the DGP model has the joint distribution,

$$p(y, \{f^l\}_{l=1}^L) = \prod_{i=1}^N p(y_i \mid f^L(f^{L-1}(\ldots f^1(x_i)))) \prod_{l=1}^L p(f^l),$$

where the first part on the right-hand side is *likelihood*, the second part is *prior*, and each layer is a *GP* $p(f^l) = \mathcal{GP}(m^l, k^l)$.

The compositional and hierarchical structure of a DGP makes it natural for a DGP to be realized using a *deep neural network*. The following shows an example code of a DGP in *Edward2* [7].

```
batch_size = 256
features, labels = load_spatial_data(batch_size)

model = tf.keras.Sequential([
    tf.keras.layers.Flatten(),
    layers.SparseGaussianProcess(units=256,num_inducing=512),
    layers.SparseGaussianProcess(units=256,num_inducing=512),
    layers.SparseGaussianProcess(units=10,num_inducing=512),
])
```

## 4.3 Deep Mixed Effects Models

*Grouped data* commonly occur in social sciences and medicine. The usual approach when dealing with multiple samples from each group is to utilize the *mixed effects models* [12]. The mixed effects are composed of two parts: the *fixed effects* and the *random effects*. The fixed effects are common in all samples and so their coefficients are called fixed. In contrast, the random effects are specific to groups and their coefficients can vary depending on the groups, which are assumed to be drawn from some unknown distributions. Using mixed effects is natural whenever data are clustered in *groups / hierarchy*.

### *Linear Mixed Effects Models*

A *linear mixed effects model* [12] can be expressed as:

$$y_i = \beta_0 + \beta_1 x_1 + \ldots + \beta_p x_p + u_{i1} z_1 + \ldots + u_{iq} z_q + \epsilon_i,$$

where $\beta := [\beta_1, \ldots, \beta_p]^T$ are the fixed effects shared over the entire sample, $u_i := [u_{i1}, \ldots, u_{ip}]^T$ are the random effects of the ith group, and $z = [z_1, \ldots, z_q]$ is a design vector for random effects. $u_i$ and $\epsilon_i$ are drawn from group-specific unknown distributions. Typically, the unknown distributions are assumed to be a zero-mean Gaussian with an unknown covariance structure:

$u_i \sim \mathcal{N}(0, \Sigma_u)$ and $\epsilon_i \sim \mathcal{N}(0, \Sigma_{\epsilon i})$.

Since the linear mixed effects models have multiple random effects from unknown distributions, no closed form solution is available. For estimation, *Expectation-Maximization* algorithms and *Markov Chain Monte Carlo* sampling are used.

### *Deep Mixed Effects Models*

A *deep mixed effects model* [12] is an extension to linear mixed effects models that uses *nonlinear mappings* between the response variable and predictor variables, which are realized using *deep neural networks*. Let $y_i = [y_{(ij)}]_{j=1}^{n_i}$ be a set of $n_i$ observations for a response variable from group i, $X_i = [x_{(ij)1}, \ldots, x_{(ij)p}]_{j=1}^{n_i}$ the corresponding predictor variables for group i, and $Z_i = [z_{(ij)1}, \ldots, z_{(ij)q}]_{j=1}^{n_i}$ the corresponding design matrix for group i. The deep mixed effects model can be written as

$y_i = \Gamma(X_i)\beta + \Gamma(Z_i)u_i,$

where $\beta$ are the coefficients for the *fixed effects* and $u_i$ are the coefficients for the *random effects* of group i. $\Gamma(.)$ is a nonlinear transformation that can be realized by a *deep neural network* of any valid architecture. Recall that

$u_i \sim \mathcal{N}(0, \Sigma_u),$

i.e., $u_i$ is drawn from a zero-mean Gaussian with an unknown covariance structure.

The deep mixed effects model thus consists of a fixed effects component, $f(x_{(ij)}) = \Gamma(x_{(ij)})\beta$, and a separate random effects component, $f'(x_{(ij)}) = \Gamma(x_{(ij)})u_i$. For regression problems, the deep mixed effects model can be written as an optimization problem by training the deep neural network $\Gamma(.)$ with the loss function

$\sum_{ij} || y_{(ij)} - f(x_{(ij)}) - f'(x_{(ij)}) ||^2.$

# 5 Summary and Conclusion

Probabilistic deep learning is deep learning that accounts for uncertainty, both model uncertainty and data uncertainty. It is based on the use of probabilistic models and deep neural networks. As background information, we discussed the probabilistic foundation of probabilistic deep learning: uncertainty, probabilistic models, and Bayesian inference.

We distinguish two approaches to probabilistic deep learning: probabilistic neural networks and deep probabilistic models. For probabilistic neural networks, we discussed Bayesian neural networks and mixture density networks; for deep probabilistic models we discussed variational autoencoders, deep Gaussian processes and deep mixed effects models. Where applicable, we included TensoFlow Probability code examples for illustration.

With the availability of libraries for probabilistic modeling and inference, such as TensorFlow Probability, it is hopeful that probabilistic deep learning will become more widely used in the near future so that uncertainty, both model uncertainty and data uncertainty, in deep learning will routinely be accounted for and quantified.

**Acknowledgement:** Thanks to my wife Hedy (期芳) for her support.

# References

[1] Daniel T. Chang, "Concept-Oriented Deep Learning: Generative Concept Representations," arXiv preprint arXiv:1811.06622 (2018).
[2] Daniel T. Chang, "Bayesian Hyperparameter Optimization with BoTorch, GPyTorch and Ax," arXiv preprint arXiv: 1912.05686 (2019).
[3] Vincent Fortuin, "Priors in Bayesian Deep Learning: A Review," arXiv preprint arXiv: 2105.06868 (2021).
[4] *TensorFlow Probability* (https://www.tensorflow.org/probability)
[5] Joshua V. Dillon, Ian Langmore, Dustin Tran, Eugene Brevdo, Srinivas Vasudevan, Dave Moore, Brian Patton, Alex Alemi, Matt Hoffman, and Rif A. Saurous, "TensorFlow Distributions," arXiv preprint arXiv:1711.10604 (2017).
[6] Dan Piponi, Dave Moore, and Joshua V. Dillon, "Joint Distributions for TensorFlow Probability," arXiv preprint arXiv: 2001.11819 (2020).
[7] Dustin Tran, Michael W. Dusenberry, Mark van der Wilk, and Danijar Hafner, "Bayesian Layers: A Module for Neural Network Uncertainty," arXiv preprint arXiv:1812.03973 (2019).
[8] Laurent Valentin Jospin, Wray Buntine, Farid Boussaid, Hamid Laga, and Mohammed Bennamoun, "Hands-on Bayesian Neural Networks - a Tutorial for Deep Learning Users", ACM Comput. Surv. 1, 1 (July 2020).
[9] Agustinus Kristiadi, Sina Daubener, and Asja Fischer, "Predictive Uncertainty Quantification with Compound Density Networks," arXiv preprint arXiv:1902.01080 (2019).
[10] Andres R. Masegosa, Rafael Cabanas, Helge Langseth, Thomas D. Nielsen, and Antonio Salmeron, "Probabilistic Models with Deep Neural Networks," arXiv preprint arXiv:1908.03442 (2019).
[11] Hugh Salimbeni, "Deep Gaussian Processes: Advances in Models and Inference," Ph.D. Thesis, Imperial College London (June 7, 2020).
[12] Y. Xiong, H. J. Kim and V. Singh, "Mixed Effects Neural Networks (MeNets) With Applications to Gaze Estimation," Proc. IEEE Conf. Comput. Vis. Pattern Recognit., pp. 7743-7752 (2019).